\pdfoutput=1
\documentclass[11pt]{article}

\usepackage[dvipsnames]{xcolor}

\usepackage[]{emnlp2021}
\usepackage{times}
\usepackage{latexsym}

\usepackage[utf8]{inputenc}

\usepackage{colortbl}
\usepackage{tikz}

\usepackage{algorithm}
\usepackage[noend]{algpseudocode}
\usepackage{graphicx}
\usepackage{hyperref}

\usepackage[T1]{fontenc}

\usepackage{comment}

\usepackage{microtype}

\begin{document}
\title{Improving Question Answering with Generation of NQ-like Questions} 

\author{Saptarashmi Bandyopadhyay \\  University of Maryland, College Park \\ \texttt{saptab1@umd.edu} 
\And Shraman Pal \\  IIT Kharagpur \\ \texttt{shramanpal@gmail.com}
\And Hao Zou \\  University of Minnesota \\ \texttt{zou00080@umn.edu }
\AND 
Abhranil Chandra  \\ IIT Kharagpur  \\ \texttt{abhranil.iitkgp@gmail.com}   
\And Jordan Boyd-Graber \\ University of Maryland, College Park\\ \texttt{jbg@umiacs.umd.edu}}

\maketitle

\begin{abstract}
Question Answering (QA) systems require a large amount of annotated data which is costly and time-consuming to gather. Converting datasets of existing QA benchmarks are challenging due to different formats and complexities. To address these issues, we propose an algorithm to automatically generate shorter questions resembling day-to-day human communication in the Natural Questions (NQ) dataset from longer trivia questions in Quizbowl (QB) dataset by leveraging conversion in style among the datasets. This provides an automated way to generate more data for our QA systems. To ensure quality as well as quantity of data, we detect and remove ill-formed questions using a neural  classifier. We demonstrate that in a low resource setting, using the generated data improves the QA performance over the baseline system on both NQ and QB data. Our algorithm improves the scalability of training data while maintaining quality of data for QA systems.
\end{abstract}

\section{Introduction}

Large-scale data collection is a challenging process in the domain of Question Answering and Information Retrieval due to the necessity of high-quality annotations which are scarce and expensive to generate. 
There are several QA datasets \cite{joshi2017triviaqa}, \cite{rajpurkar2016squad}, \cite{yang2018hotpotqa}, \cite{kwiatkowski-etal-2019-natural}, \cite{rodriguez2021quizbowl} with significantly different structure and complexity. Large quantities of high quality data are effective in training a more efficacious Machine Learning system. 

In this paper, we focus on converting questions normally spoken in a trivia competition to questions resembling day-to-day human communication. Trivia questions have multiple lines, consist of multiple hints as standalone sentences to an answer, and players can buzz on any sentence in the question to give an answer. In contrast, questions used in daily human communication are shorter (often a single line). We propose an algorithm to generate multiple short natural questions from every long trivia question by converting each of the sentences with multiple hints to several shorter questions. We also add a BERT \cite{devlin2019bert} based quality control method to filter out the ill-formed questions and retain the well-formed questions. We show that our algorithm for generating questions improves the performance of two question answering (QA) systems in a low resource setting. We also demonstrate that concatenation of original natural questions with generated questions improves the QA system performance. Finally, we prove that by using such a method to generate synthetic data, we can achieve higher scores than a system that uses only NQ data. 

\section{Dataset and Data Extraction}

We use two popular datasets- the Quizbowl \cite{rodriguez2021quizbowl}, henceforth referred to as QB dataset, and NQ-Open dataset \cite{lee-etal-2019-latent}, derived from Natural Questions \cite{kwiatkowski-etal-2019-natural}. QB has a total of 119247 question/answer samples and NQ has 91434 total question/answer samples. For the NQ dataset, we use the same 1800 dev and 1769 test question/answer splits as used in the EfficientQA Competition \cite{min2021neurips}.

As our task involves transforming QB questions to NQ-like questions, we extract pairs of questions that are semantically similar. We first extract every possible question-question pair with the same answer using string matching resulting in 95651 question-question pairs. From this parallel corpus, we extract the last sentence of the QB questions and pass them through a pre-trained Sentence-BERT \cite{reimers2019sentencebert} model
along with the corresponding NQ question. We take the cosine similarity between the [CLS] embedding to find pairs that are semantically equivalent by setting the threshold to 0.5. From this we extract 19439 question-question pairs having moderate semantic equivalence. We also use the same index from the last sentence of QB questions paired with NQ questions to retrieve corresponding QB full questions in paragraph form. Out of the extracted corpora, we create a smaller dataset of the last sentence of every QB question (and corresponding QB full paragraph) paired with NQ questions for our low resource setting. This paired dataset has a total of 1218 training samples, 93 validation samples and 563 test samples which are semantically similar. We outline the statistics of the baseline and generated datasets in Table \ref{statDataQA}.


\section{Methods to generate NQ-like Questions}\label{sec:methods}
We use the following NLP techniques to generate NQ-like questions from our QB dataset
\begin{itemize}
    \item Tokenization
    \item Coreference resolution
    \item Parse tree output
    \item Bag of words based question generation
\end{itemize}
We outline our methods in Algorithm \ref{alg:qgen}. At first, we tokenize the long paragraph of a QB question into individual sentences with hints to an answer providing sufficient context. Using every sentence annotated by the question id was not sufficient due to annotation errors of sentence delimiters like a `.'. 
\begin{itemize}
   \item \textbf{\underline{Initial}:}  ... and "k." For 10 points, ...
   \item \textbf{\underline{Tokenized}:} 
          \begin{itemize}
              \item ... and " k. "
              \item For 10 points , ... 
          \end{itemize}
\end{itemize}
However, each sentence contains 2-3 clues about the answer whereas NQ questions generally are in 1 sentence with 1 clue. To resolve this issue, we use Coreference Resolution on every sentence to obtain clusters of nouns and pronouns referring to those nouns \cite{kirstain2021coreference}, and Parse Trees.

We then split the sentences to get parts of a sentence with similar syntactic structure using Parse Tree based on ADVCL (adverbial clause) and CC (conj) tags. Any pronoun that is present in a split is replaced with the noun from the clusters found using Coreference Resolution. After breaking into splits, we run a check on the number of words in every split and append sentences with smaller number of words (less than 8) back into the original sentence. We finally clean the split sentences of trailing punctuation marks and words that should not be at the sentence ending like `and', `but' etc. This gives us sentences or phrases that usually have one clue similar to NQ.

To make the outputs question-like, we use a bag of words approach by replacing words like `this' with `which', `it' with `what' for all but the last sentence. The last sentence in QB has a specific syntactic structure where the sentence contains, `For x points , name this' followed by the identifier to the answer. We replace this whole phrase according to whatever comes after. For example, 
\begin{itemize}
     \item`name this author' : `who is the author'
     \item`name this 1985 event' : `which is the 1985 event'
     \item`name this phenomenon' : `what is the phenomenon'
    \end{itemize}

\begin{algorithm}[!h]
\caption{NQ-like Question Generation}\label{alg:qgen}
\begin{algorithmic}[1]
\Procedure{NQlikegen}{$sentences$}

\State $Individual(sentences)$  \Comment{split full QB question into single sentences}
\State $Pre\_processing(sentences)$ \Comment{remove punctuations, conjunctions or any other word not expected at the end of the sentence.}
\State $Coreference\_clusters(sentences)$ \Comment{forms clusters like [“this country”, “its”]}
\State $Parse\_Tree(sentences)$ \Comment{Split the conjoined clues based on similar syntactic structure}


\If{$Noun , Pronoun$ not in $same\_parse\_split$}
\State $Replace$ \Comment{replace the pronoun in the other split with the NP}
\EndIf
\For{$QB\_last\_sentences$}
\State $String\_Replacement(last\_sentence)$ \Comment{Replace ‘name this’ with the annotated ‘wh’ question by NOUNS vocabulary formed}
\EndFor
\For{$QB\_non\_last\_sentences$}
\State $Bag\_of\_Words$ \Comment{use frequency based vocabulary table to replace “this” to “which”}
\EndFor

\State \textbf{return} $nq\_like\_questions$
\EndProcedure
\end{algorithmic}
\end{algorithm}

To achieve this, we extract noun phrases following `name this' from  1000 samples and form a vocabulary to manually annotate them with `who is the', `which is the', `what is the', and their plural counterparts. This gives us questions that resemble NQ questions.  






\begin{table*}[!h]
\centering
\small
\begin{tabular}{|l|l|l|l|l|l|l|l|}
\hline
\textbf{System Description}  & \textbf{Data Size} & \textbf{Mean} & \textbf{Median} & \textbf{Mode} \\ \hline

QB system with full questions & 1874        & 4.39	& 4.0 &	4.0	   \\ \hline
Generated NQ-like full questions & 11062   & 5.9	& 6.0 &	6.0	   \\ \hline
Filtered generated NQ-like full questions  & 2075             & 1.7	& 1.0 &	1.0	   \\ \hline
\hline

QB system with full questions & 119247              & 6.2	& 6.0 &	6.0	   \\ \hline
Generated NQ-like full questions & 772456             & 6.5	& 6.0 &	6.0	   \\ 
\hline

\end{tabular}
\caption{Statistics of number of sentences per Quizbowl question}
\label{statDataQA}
\end{table*}

\begin{figure*}[!h]    
\centering
    \includegraphics[width=1.0\textwidth,height=1.1\textwidth]{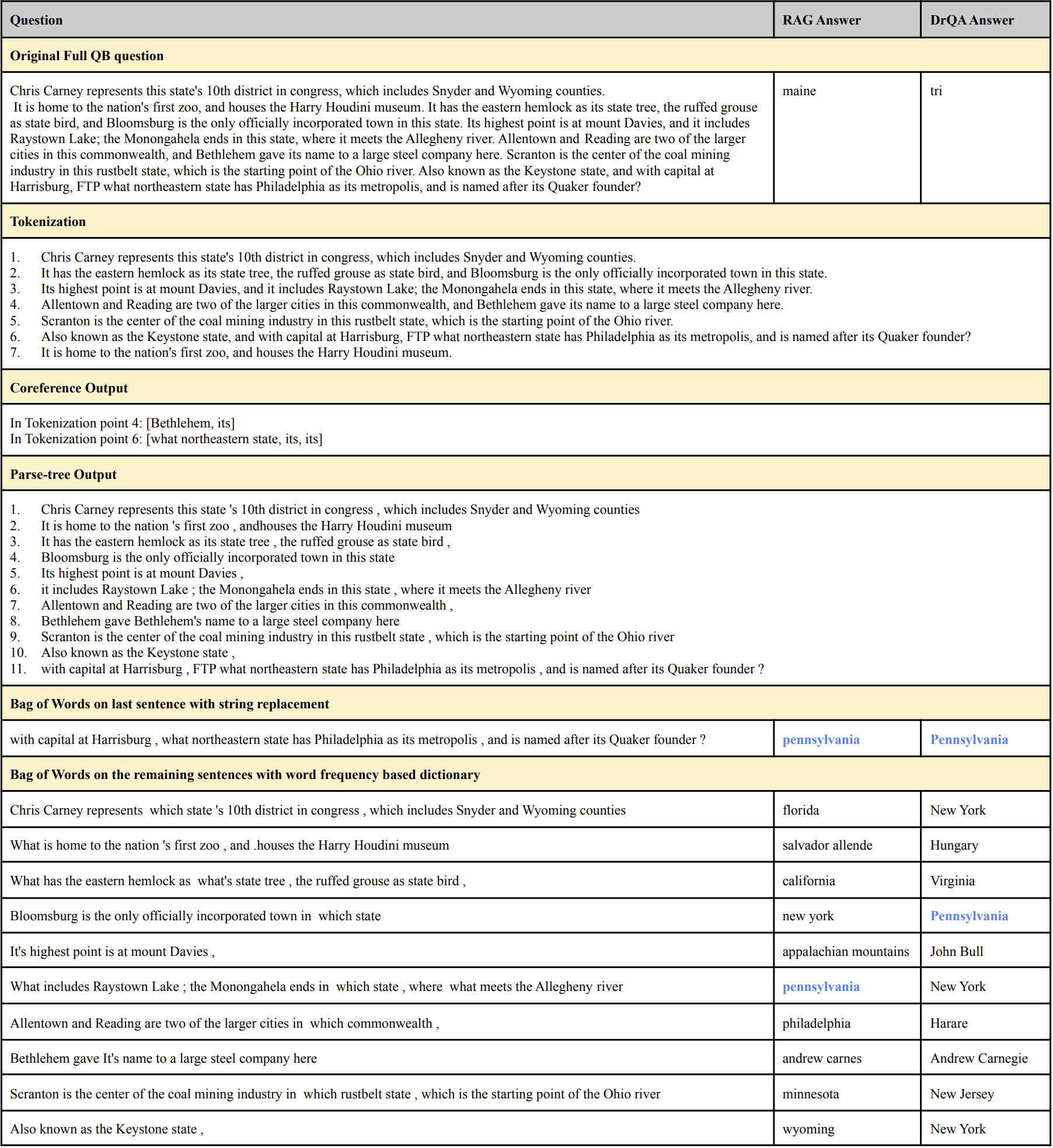}
    \caption{Generation of NQ-like Questions from the original QB Questions along with answers of the QA systems (Gold Answer: {\textcolor{Periwinkle}{pennsylvania}})} 
    \label{fig:QA Transformation Example}
\end{figure*}

\begin{table*}[!ht]
\centering
\small
\begin{tabular}{|l|l|l|l|l|l|l|l|l|}
\hline
\textbf{System} & \textbf{Total samples} & \textbf{Accuracy} & \textbf{Precision} & \textbf{Recall} & \textbf{F1} & \textbf{SacreBLEU}\\ \hline

NQ & 1874   & 0.490	& 0.550 &	0.540	& 0.540 &	29.57    \\ \hline
NQ + NQ-like from QB full &   9140          & 0.390	& 0.420 &	0.410	& 0.420                                                                                                                          &	27.95    \\ \hline
NQ + Quality controlled NQ-like from QB full  & 3319   & 0.467	& 0.505 &	0.503	& 0.502 &	34.03    \\ \hline

{\color[HTML]{212121} NQ + QB full}  & {\color[HTML]{212121}3092}      & {\color[HTML]{212121}0.465}             & {\color[HTML]{212121} 0.503}        & {\color[HTML]{212121} 0.502}          & {\color[HTML]{212121} 0.499}           & {\color[HTML]{212121} 27.62}              \\ \hline
{\color[HTML]{212121} NQ + QB last sentence} & {\color[HTML]{212121} 3092} & {\color[HTML]{212121} 0.516}             & {\color[HTML]{212121} 0.558}        & {\color[HTML]{212121} 0.560}          & {\color[HTML]{212121} 0.556}           & {\color[HTML]{212121} 37.37}               \\ \hline

{\color[HTML]{212121} NQ + NQ-like from QB last sentence} & {\color[HTML]{212121} 3259}      & {\color[HTML]{212121} 0.500}             & {\color[HTML]{212121} }0.557        & {\color[HTML]{212121} } 0.556         & {\color[HTML]{212121} }0.553           & {\color[HTML]{212121} 34.98}              \\ \hline
{\color[HTML]{212121} NQ + Quality controlled NQ-like from QB last sentence} & {\color[HTML]{212121}2967} & {\color[HTML]{212121}\textbf{0.518}}             & {\color[HTML]{212121} \textbf{0.567}}        & {\color[HTML]{212121} \textbf{0.564}}          & {\color[HTML]{212121} \textbf{0.562}}           & {\color[HTML]{212121} \textbf{37.46}}               \\ \hline
\hline
{\color[HTML]{212121} QB full} & {\color[HTML]{212121} 1874} & {\color[HTML]{212121} 0.232}             & {\color[HTML]{212121} 0.261}        & {\color[HTML]{212121} 0.259}          & {\color[HTML]{212121} 0.259}           & {\color[HTML]{212121} 12.17}               \\ \hline
{\color[HTML]{212121} NQ-like from QB full} & {\color[HTML]{212121}11062} & {\color[HTML]{212121} 0.171}             & {\color[HTML]{212121}0.188}        & {\color[HTML]{212121} 0.188}          & {\color[HTML]{212121} 0.188}           & {\color[HTML]{212121} 12.45}               \\ \hline
{\color[HTML]{212121} Quality controlled NQ-like from QB full} & {\color[HTML]{212121} 2075} & {\color[HTML]{212121} 0.176}             & {\color[HTML]{212121} 0.192}     & {\color[HTML]{212121} 0.191}   & {\color[HTML]{212121} 0.191}     & {\color[HTML]{212121} 14.11}               \\ \hline
{\color[HTML]{212121} QB last sentence} & {\color[HTML]{212121} 1874} & {\color[HTML]{212121} 0.44}             & {\color[HTML]{212121} \textbf{0.495}} & {\color[HTML]{212121} \textbf{0.488}} & {\color[HTML]{212121} \textbf{0.488}}  & {\color[HTML]{212121} 22.18}  \\ \hline
{\color[HTML]{212121} NQ-like from QB last sentence} & {\color[HTML]{212121} 2112} & {\color[HTML]{212121} 0.421}             & {\color[HTML]{212121} 0.454} & {\color[HTML]{212121} 0.456} & {\color[HTML]{212121} 0.453}  & {\color[HTML]{212121} 34.49}  \\ \hline
{\color[HTML]{212121} Quality controlled NQ-like from QB last sentence}& {\color[HTML]{212121} 1632} & {\color[HTML]{212121} \textbf{0.441}}             & {\color[HTML]{212121} \textbf{0.495}} & {\color[HTML]{212121} \textbf{0.488}} & {\color[HTML]{212121} \textbf{0.488}}  & {\color[HTML]{212121} \textbf{38.25}}  \\ \hline
\end{tabular}
\caption{Question Answering Metric values with RAG system}
\label{RAGQA}
\end{table*}

\begin{table*}[!ht]
\centering
\small
\begin{tabular}{|l|l|l|l|l|l|l|l|l|}
\hline
\textbf{System} & \textbf{Total samples} & \textbf{Accuracy} & \textbf{Precision} & \textbf{Recall} & \textbf{F1} & \textbf{SacreBLEU}\\ \hline

NQ  & 1874   & 0.211	& 0.243 &	0.242	& 0.241 &	8.74    \\ \hline
NQ + NQ-like from QB full &   9140           & 0.243	&0.271  &0.270	&0.269  & 10.49    \\ \hline
NQ + Quality controlled NQ-like from QB full   & 3319   & 0.227	& 0.256 &	0.254	& 0.254 &	\textbf{11.01}    \\ \hline
{\color[HTML]{212121} NQ + QB full}  & {\color[HTML]{212121}3092}      & {\color[HTML]{212121}0.229}             & {\color[HTML]{212121} 0.257}        & {\color[HTML]{212121} 0.258}          & {\color[HTML]{212121} 0.256}           & {\color[HTML]{212121} 10.97}              \\ \hline
{\color[HTML]{212121} NQ + QB last sentence} & {\color[HTML]{212121} 3092} & \textbf{{\color[HTML]{212121}0.257}}             & \textbf{{\color[HTML]{212121} 0.291}}        & \textbf{{\color[HTML]{212121} 0.295}}          & \textbf{{\color[HTML]{212121} 0.290}}           & {\color[HTML]{212121} 9.65}              \\ \hline
{\color[HTML]{212121} NQ + NQ-like from QB last sentence} & {\color[HTML]{212121} 3259}      & {\color[HTML]{212121} 0.253}             & {\color[HTML]{212121} }0.281        & {\color[HTML]{212121} } 0.276         & {\color[HTML]{212121} }0.277           & {\color[HTML]{212121} 10.20}              \\ \hline
{\color[HTML]{212121} NQ + Quality controlled NQ-like from QB last sentence} & {\color[HTML]{212121}2967} & {\color[HTML]{212121}0.234}             & {\color[HTML]{212121} 0.268}        & {\color[HTML]{212121} 0.267}          & {\color[HTML]{212121} 0.266}           & {\color[HTML]{212121} 8.91}               \\ \hline
\hline
{\color[HTML]{212121} QB full} & {\color[HTML]{212121} 1874} & {\color[HTML]{212121} 0.191}             & {\color[HTML]{212121} 0.251}        & {\color[HTML]{212121} 0.228}          & {\color[HTML]{212121} 0.233}           & {\color[HTML]{212121} 7.17}               \\ \hline
{\color[HTML]{212121} NQ-like from QB full} & {\color[HTML]{212121}11062} & {\color[HTML]{212121} 0.027}             & {\color[HTML]{212121}0.085}        & {\color[HTML]{212121} 0.085}          & {\color[HTML]{212121} 0.085}           & {\color[HTML]{212121} 2.23}               \\ \hline
{\color[HTML]{212121} Quality controlled NQ-like from QB full} & {\color[HTML]{212121} 2075} & {\color[HTML]{212121} 0.197}             & {\color[HTML]{212121} 0.224}     & {\color[HTML]{212121} 0.218}   & {\color[HTML]{212121} 0.219}     & {\color[HTML]{212121} 8.71}               \\ \hline
{\color[HTML]{212121} QB last sentence } & {\color[HTML]{212121} 1874} & {\color[HTML]{212121} 0.328}             & {\color[HTML]{212121} 0.394} & {\color[HTML]{212121} 0.384} & {\color[HTML]{212121} 0.383}  & {\color[HTML]{212121} 9.66}  \\ \hline
{\color[HTML]{212121} NQ-like from QB last sentence} & {\color[HTML]{212121} 2112} & {\color[HTML]{212121} 0.310}             & {\color[HTML]{212121} 0.363} & {\color[HTML]{212121} 0.358} & {\color[HTML]{212121} 0.357}  & \textbf{{\color[HTML]{212121} 25.40}}  \\ \hline
{\color[HTML]{212121} Quality controlled NQ-like from QB last sentence}& {\color[HTML]{212121} 1632} & \textbf{{\color[HTML]{212121} 0.367}}             & \textbf{{\color[HTML]{212121} 0.428}} & \textbf{{\color[HTML]{212121} 0.429}} & \textbf{{\color[HTML]{212121} 0.424}}  & {\color[HTML]{212121} 23.16 }  \\ \hline
\end{tabular}
\caption{Question Answering Metric values with DrQA system}
\label{DrQA}
\end{table*}

\subsection{Quality Control}\label{sec:qualitycontrol}
We trained a BERT based \cite{devlin2019bert} classifier using the pretrained bert-base-uncased model for detecting well-formed questions and improving the data quality. We use the dataset from \cite{FaruquiDas2018} which has questions annotated with scores from 0 to 1 signifying how well the questions are structured. With the help of this classifier, we generate the scores for our training set (7266 samples) of generated NQ like questions. We only take the questions with a score greater than 0.5 to maintain a good balance between quality and quantity. This provides us with 1445 synthetic good quality samples.
We add these good-quality NQ-like questions to our original NQ training set thereby increasing the number of samples to 2663 QA pairs. In the example shown in Figure \ref{fig:QA Transformation Example}, the well-formed questions detected by the classification model are:
\begin{itemize}
\item Chris Carney represents  which state 's 10th district in congress , which includes Snyder and Wyoming counties
\item What includes Raystown Lake ; the Monongahela ends in  which state , where  what meets the Allegheny river
\item With capital at Harrisburg , what northeastern state has Philadelphia as its metropolis , and is named after its Quaker founder ?
\end{itemize}

\section {Experiments}
\label{sec:experiments}
We use a smaller subset (1874 samples) from the total paired samples to form our paired dataset containing NQ and QB questions. The train, dev, and test splits are 1218, 93, and 563 respectively. We provide generated samples sizes using our algorithm and the scores of two QA systems; DrQA and RAG which trained on these varying experiments. 

\subsection{Generated Data}\label{sec:gendata} 

We originally possess 1874 total samples [\textit{1218 train, 93 dev, 563 test}] for both the NQ dataset and QB full dataset consisting of QB questions having multiple sentences. We extract only the last sentence from the QB questions (QB last sentence) with the same split [\textit{1218 train, 93 dev, 563 test}]. We generate NQ-like questions from QB questions with all sentences (NQ-like from QB full) [\textit{7266 train, 517 dev, 3279 test}] and questions in the last sentence of QB questions (NQ-like from QB last sentence) [\textit{1385 train, 105 dev, 622 test}], as outlined in Algorithm \ref{alg:qgen}. We then extract only the well-formed questions by utilizing a quality control classifier as outlined in Section \ref{sec:qualitycontrol}. From this we obtain the filtered NQ-like questions from the QB questions with multiple sentences (Quality controlled NQ-like from QB full) dataset \textit{[1445 train, 113 dev, 517 test]}, and the filtered NQ-like questions from the last sentence of the QB questions (Quality controlled NQ-like from QB last sentence) \textit{[1093 train, 74 dev, 465 test]}. The NQ dataset is concatenated with the 6 different types of datasets outlined above. Assuming a generic \textit{[Dataset Name]} for the 6 datasets, we name the augmented system like (NQ + \textit{[Dataset Name]}) shown in Tables \ref{RAGQA} and \ref{DrQA}.

\subsection{Question Answering Systems}
After reviewing several QA systems \cite{zhu2021retrieving}, we finally used the Retrieval Augmented Generation (RAG) \cite{lewis2021retrievalaugmented} and DrQA \cite{chen2017reading} as our baseline QA systems. We used a small subset of the Wikipedia dump as our retrieval dump for the low resource setting. 

\paragraph{\textbf{RAG QA System}}
    We retrieve $5$ documents using the RAG retriever and use a batch size of $2$ for our training. We train our systems for $5$ epochs by using AdaGrad \cite{duchi2011adaptive} as the optimizer with a learning rate of $1e^{-4}$.

\paragraph{\textbf{DrQA System}}
    We retrieve $10$ documents for gold passage computation. We use SGD as the optimizer with $0.1$ learning rate, and pre-trained word vector GloVe with 840 billion tokens and 300 dimensional vector embeddings \cite{pennington2014glove}. 3-layer bidirectional long short-term memory network (LSTM) with $128$ hidden units is used for both paragraph and question encoding. The numbers of document layers and answer layers are both set to $3$. We train all our systems for $50$ epochs with batch size $64$ and max length $30$.

{
\section{Results} 
\label{sec:results} 
Our research work demonstrates a clear improvement in QA performance for RAG and DrQA systems as shown in Tables \ref{RAGQA} and \ref{DrQA}. We see an improvement in QA evaluation scores when trivia questions with multiple sentences in QB dataset can be used to automatically generate NQ-like questions which are then concatenated with the original NQ questions. For the RAG QA system, the baseline score was improved by 3 of our proposed systems, out of which generated outputs from the last sentence concatenated with original NQ data has seen the most success with an improvement of 2.8 points in accuracy and an increase of 1.7, 2.4, 2.2 and 7.89 points on Precision, Recall, F1 measure, and BLEU respectively. The performance of the QA system trained on NQ-like filtered data concatenated with the NQ data is also better than the system based on the concatenation of QB dataset with NQ dataset. For DrQA, all of our proposed systems have done better than the baseline with NQ data and the system with concatenated NQ and QB data. The generated outputs from the last sentence of the QB concatenated with NQ have shown an improvement of 4.6 points in accuracy and 4.8, 5.3, 4.9, 0.91 points in Precision, Recall, F1 measure, and BLEU respectively on the DrQA system. Most of our systems exhibit an increase in BLEU score over the baseline.  We also observe that the performance of RAG and DrQA systems trained on only NQ like data generated from QB dataset is better than the baseline systems on QB dataset. 

This is understandable as the last sentence usually has the easiest clue. On top of which, it has a regular structure. These characteristics make the last sentence from the QB  more semantically aligned with NQ-like data after conversion using our algorithm.

Our proposed algorithm as outlined in Algorithm \ref{alg:qgen} and quality filtering technique (1093 extra samples) as mentioned in Section \ref{sec:qualitycontrol} can be effective in a low resource setting with a lack of sufficiently well annotated QA data. This also helps in scaling the training data by converting multiple datasets automatically.


A broad example of NQ-like question generation and subsequent answer generation has been provided in Figure \ref{fig:QA Transformation Example}.

}

%


\section{Conclusion and Future Work}
\label{sec:conclusion and future work} 
 We clearly observe from the results that adding filtered NQ-like questions from the QB data has given a boost over using only NQ questions. In finer detail, we observe that questions from the last sentence of the QB are of higher quality than from intermediate sentences and therefore provide a higher boost to performance even with less samples. Even by simply adding questions generated from last sentence, we increase the exact match accuracy by nearly 2 points. We also observe that the BLEU score of answers generated from quality controlled NQ like system is 16 points more than the BLEU score of the baseline QB system for the RAG system and by 13 points for the DrQA system.  This shows that our algorithm to generate NQ-like questions has been effective in improving the quality of the training dataset.

We are working to extend this system from the smaller datasets to the entire 119247 QB and 91494 NQ samples by using the generated 772456 unfiltered NQ-like questions from the entire QB data, retain the well-formed questions and improve the performance over the baseline system trained on NQ dataset.

We observe that QB trivia questions are in passive voice while NQ questions are in active voice. So Neural Machine Translation can help in converting the style of QB questions. Our dataset pairing method automatically filters out adversarial samples and improves the quality of the generated NQ-like corpus. This opens the door towards a possible research area where multiple non-adversarial datasets can be used to filter the NQ dataset from adversarial samples. We plan on improving the filtering process by generating our own annotated dataset regarding well-formed and ill-formed questions in order to detect well formed questions and merge ill formed questions with better quality questions.

Specifically, through our manual transformation for NQ-like questions, we are specifying three different answer types like WHO/WHICH/WHAT, which might be useful to propagate to intermediate parse outputs during NQ-like question generation in order to improve answer generation.

Other information like difficulty and year associated with QB questions can be leveraged to improve quality or filtering and can even be used to predict extra information on NQ questions.

Further work can also be done on augmenting other datasets to NQ-like questions to generate more synthetic samples which may improve question answering performance.

\bibliography{anthology,custom}
\bibliographystyle{acl_natbib}




\end{document}